\pdfoutput=1

\documentclass[11pt]{article}

\usepackage[final]{acl}

\usepackage{times}
\usepackage{latexsym}

\usepackage[T1]{fontenc}

\usepackage[utf8]{inputenc}

\usepackage{microtype}

\usepackage{inconsolata}

\usepackage{graphicx}

\usepackage{multirow}
\usepackage{booktabs}
\usepackage{pifont}
\usepackage{graphicx} 

\newcommand{\rot}[1]{\rotatebox{90}{\footnotesize #1}}

\usepackage{adjustbox}
\usepackage{rotating}
\usepackage{threeparttable}

\usepackage{acro}

\usepackage{wasysym}

\DeclareAcronym{cd}{
  short = CD ,
  long  = cognitive distortion ,
  short-plural = s ,
}
\DeclareAcronym{cbt}{
  short = CBT ,
  long  = cognitive behavioural therapy 
}
\DeclareAcronym{llm}{
  short = LLM ,
  long  = large language model 
}
\DeclareAcronym{nlp}{
  short = NLP ,
  long  = natural language processing 
}
\DeclareAcronym{iaa}{
  short = IAA ,
  long  = inter-annotator agreement
}

\title{A Survey of Cognitive Distortion Detection and Classification in NLP}

\author{
Archie Sage,
Jeroen Keppens,
Helen Yannakoudakis \\
Department of Informatics, King’s College London \\
\texttt{\{archie.sage, jeroen.keppens, helen.yannakoudakis\}@kcl.ac.uk}
}

\begin{document}
\maketitle

\begin{abstract}

As interest grows in applying \ac{nlp} techniques to mental health, an expanding body of work explores the automatic detection and classification of \acp{cd}. \acp{cd} are habitual patterns of negatively biased or flawed thinking that distort how people perceive events, judge themselves, and react to the world. Identifying and addressing them is a central goal of therapy. Despite this momentum, the field remains fragmented, with inconsistencies in \ac{cd} taxonomies, task formulations, and evaluation practices limiting comparability across studies. This survey presents the first comprehensive review of 38 studies spanning two decades, mapping how \acp{cd} have been implemented in computational research and evaluating the methods applied. We provide a consolidated \ac{cd} taxonomy reference, summarise common task setups, and highlight persistent challenges to support more coherent and reproducible research. Alongside our review, we introduce practical resources, including curated evaluation metrics from surveyed papers, a standardised datasheet template, and an ethics flowchart, available online.\footnote{\url{https://github.com/archiesage/cognitive-distortion-nlp-survey}}

\end{abstract}
\acresetall 

\section{Introduction}
\label{sec:introduction}

\Acp{cd} are habitual patterns of negatively biased or logically flawed thinking that distort how people perceive events, judge themselves, and react to the world around them. These distortions play a central role in emotional distress and are a core target of evidence-based psychological interventions such as \ac{cbt} \citep{beck_1963_thinking, burns_1999_feeling}.

Common examples\footnote{A complete list of \ac{cd} definitions, examples, and synonyms is provided in Table~\ref{tab:cd_definitions}, with additional psychological background in Appendix~\ref{sec:psychology-foundation-appendix}.} include \textit{Catastrophising} (`I let them down, so they'll never trust me again'), \textit{Mind Reading} (`They haven’t replied, so they must be angry at me'), and \textit{All or Nothing Thinking} (`If I don't get this right the first time, I’m a complete failure').
These patterns often appear intuitive or harmless at first, but they have been shown to maintain and exacerbate conditions like depression, anxiety, and post-traumatic stress disorder. In therapeutic settings, recognising and \textit{reframing} such distortions is a core goal of \ac{cbt}. Crucially, these distortions are primarily expressed through language, making them well-suited to computational modelling. Recent work in \ac{nlp} has begun to explore the automatic detection and classification of \acp{cd}, with applications ranging from clinical decision support tools to mental health chatbots, journaling tools, and triage systems. Studies have shown that incorporating \ac{cd}-level features can improve outcomes in related tasks such as depression detection \citep{wang_2023_c2d2}, complementing more traditional sentiment or topic based approaches. By identifying distorted cognitive patterns in everyday text, \ac{nlp} systems may support more timely, personalised, and psychologically-informed interventions. 

Despite rapid growth in the field, the literature remains fragmented. Computational approaches use inconsistent taxonomies for defining \acp{cd}, making it difficult to compare findings across studies. Task formulations (e.g., detection vs classification, single-label vs multi-label) vary widely, often reflecting implicit assumptions that shape evaluation and outcomes. Benchmarks are scarce, metrics inconsistently applied, and variations in domain and dataset usage further complicate comparisons, making it hard to pinpoint gaps or establish best practices.

{
\newcommand{\dir}[1]{\textcolor{#1}{$\CIRCLE$}} 
\newcommand{\inh}[1]{\textcolor{#1}{$\LEFTcircle$}} 
\newcommand{\con}[1]{\textcolor{#1}{$\Circle$}} 

\renewcommand{\dir}[1]{\textcolor{black}{$\CIRCLE$}}      
\renewcommand{\inh}[1]{\textcolor{black}{$\LEFTcircle$}}  
\renewcommand{\con}[1]{\textcolor{black}{$\Circle$}}      


\begin{table*}[!t]
  \small
  \centering
  \setlength{\tabcolsep}{2.5pt}
  \renewcommand{\arraystretch}{1.1}
  \begin{adjustbox}{width=\textwidth}
  \begin{tabular}{@{} l l *{35}{c}@{\hspace{8pt}} r @{}}
    \toprule
    \textbf{Code} & \textbf{Cognitive Distortion}
        & \rot{\textsc{\textbf{TherapistQA}}}
        & \rot{\citet{shreevastava_2021_detecting}}
        & \rot{\citet{chen_2023_empowering}}
        & \rot{\citet{lim_2024_erd}}
        & \rot{\citet{pico_2025_comparative}}
        & \rot{\citet{zhang_2025_cbt}}
        & \rot{\citet{babacan_2025_creating}}
        & \rot{\citet{varadarajan_2025_linking}}

        & \rot{\textsc{\textbf{Text Intervention}}}
        & \rot{\citet{lybarger_2022_identifying}}
        & \rot{\citet{ding_2022_improving}}
        & \rot{\citet{tauscher_2023_automated}}

        & \rot{\textsc{\textbf{Thinking Trap}}}
        & \rot{\citet{sharma_2023_cognitive}}
        & \rot{\citet{agarwal_2025_exploratory}}

        & \rot{\textsc{\textbf{C-Journal}}}
        & \rot{\citet{elsharawi_2024_cjournal}}
        & \rot{\citet{rasmy_2025_enhanced}}

        & \rot{\textsc{\textbf{CDS}}}
        & \rot{\citet{bathina_2021_depressed}}
        & \rot{\citet{lalk_2024_depression}}

        & \rot{\textsc{\textbf{Others}}}
        & \rot{\citet{wiemer_2004_automatic}}
        & \rot{\citet{xing_2017_cnn}}
        & \rot{\citet{rojas-barahona_2018_deep-learning}}
        & \rot{\citet{shickel_2020_mental-health-text}}
        & \rot{\citet{lee_2021_micromodels}}
        & \rot{\citet{mostafa_2021_automatic}}
        & \rot{\citet{alhaj_2022_improving}}
        & \rot{\citet{wang_2023_c2d2}}
        & \rot{\citet{wang_2023_cognitive}}
        & \rot{\citet{maddela-ung_2023_training}}
        & \rot{\citet{lin_2024_detection}}
        & \rot{\citet{qi_2024_supervised}}
        & \rot{\citet{kim_2025_koacd}}
      & \textbf{\%}\textsuperscript{†} \\
    \midrule
    \multicolumn{38}{@{}l}{\textbf{Widely Adopted:} Frequently seen in \ac{nlp}, typically with clearer semantic distinctions, and recommended as a focus for future research.} \\
    \midrule
    OVG & Overgeneralisation 
      & & \dir{purple}  & \inh{purple}  & \inh{purple}  & \inh{purple}  & \inh{purple}
      & \inh{purple}  & \inh{purple}  & & \dir{red} & \inh{red}  & \inh{red}
      & & \dir{blue}    & \inh{blue}  & & \dir{orange}    & \inh{orange} & & \dir{cyan}
      & \inh{cyan}  & & \dir{black} & \dir{black} & \dir{black} & \dir{black}
      & \con{black}   & \dir{black}  & \dir{black} & \dir{black} & \dir{black}
      & \dir{black} & \con{black} & \dir{black} & \dir{black}  & 100\% \\
    SHD & Should Statements
      & 
      & \dir{purple} & \inh{purple} & \inh{purple} & \inh{purple} & \inh{purple}
      & \inh{purple} & \inh{purple}

      & 
      & \dir{red} & \inh{red} & \inh{red}

      & 
      & \dir{blue} & \inh{blue}

      & 
      & \dir{orange} & \inh{orange}

      & 
      & \dir{cyan} & \inh{cyan}

      & 
      & \dir{black} & \dir{black} & \dir{black} & \dir{black}
      & 
      & \dir{black} & \dir{black}
      & 
      & \dir{black}
      & \dir{black} & \con{black} & \dir{black} & \dir{black}

      & 93\% \\
    LBL & Labelling
      & 
      & \dir{purple} & \inh{purple} & \inh{purple} & \inh{purple} & \inh{purple}
      & \inh{purple} & \inh{purple}

      & 
      &  &  &

      & 
      & \dir{blue} & \inh{blue}

      & 
      & \dir{orange} & \inh{orange}

      & 
      & \dir{cyan} & \inh{cyan}

      & 
      & \dir{black} & \dir{black} & \dir{black} & \dir{black}
      & \dir{black}
      &  & \dir{black}
      & \dir{black}
      & \dir{black}
      & \dir{black} & \con{black} & \dir{black} & \dir{black}

      & 86\% \\
    AON & All or Nothing Thinking
      & 
      & \dir{purple} & \inh{purple} & \inh{purple} & \inh{purple} & \inh{purple}
      & \inh{purple} & \inh{purple}

      & 
      &  &  &

      & 
      & \dir{blue} & \inh{blue}

      & 
      & \dir{orange} & \inh{orange}

      & 
      & \dir{cyan} & \inh{cyan}

      & 
      & \dir{black} & \dir{black} & \dir{black} & \dir{black}
      & \dir{black}
      &  & 
      & \dir{black}
      & \dir{black}
      & \dir{black} & \con{black} & \dir{black} & \dir{black}

      & 83\% \\
    EMR & Emotional Reasoning
      & 
      & \dir{purple} & \inh{purple} & \inh{purple} & \inh{purple} & \inh{purple}
      & \inh{purple} & \inh{purple}

      & 
      &  &  &

      & 
      & \dir{blue} & \inh{blue}

      & 
      & \dir{orange} & \inh{orange}

      & 
      & \dir{cyan} & \inh{cyan}

      & 
      & \dir{black} & \dir{black} & \dir{black} & \dir{black}
      & 
      &  & \dir{black}
      & \dir{black}
      & \dir{black}
      & & \con{black} & \dir{black} & \dir{black}

      & 79\%     \\
    PRS & Personalisation
      & 
      & \dir{purple} & \inh{purple} & \inh{purple} & \inh{purple} & \inh{purple}
      & \inh{purple} & \inh{purple}

      & 
      &  &  &

      & 
      & \dir{blue} & \inh{blue}

      & 
      & \dir{orange} & \inh{orange}

      & 
      & \dir{cyan} & \inh{cyan}

      & 
      & \dir{black} & \dir{black} & \dir{black} & \dir{black}
      & 
      &  & 
      & \dir{black}
      & \dir{black}
      & \dir{black} & \con{black} & \dir{black} & \dir{black}

      & 79\% \\
    MFL & Mental Filter
      & 
      & \dir{purple} & \inh{purple} & \inh{purple} & \inh{purple} & \inh{purple}
      & \inh{purple} & \inh{purple}

      & 
      & \dir{red} & \inh{red} & \inh{red}

      & 
      &  & 

      & 
      & \dir{orange} & \inh{orange}

      & 
      & \dir{cyan} & \inh{cyan}

      & 
      &  & \dir{black} & \dir{black} & \dir{black}
      & 
      &  & 
      & 
      & \dir{black}
      & \dir{black} & \con{black} & \dir{black} & \dir{black}

      & 76\% \\
    MDR & Mind Reading
      & 
      & \dir{purple} & \inh{purple} & \inh{purple} & \inh{purple} & \inh{purple}
      & \inh{purple} & \inh{purple}

      & 
      &  &  &

      & 
      &  \dir{blue}& \inh{blue} 

      & 
      &  & 

      & 
      & \dir{cyan} & \inh{cyan}

      & 
      &  \dir{black} & \con{black} & \dir{black} & \dir{black}
      & 
      &  & 
      & \dir{black}
      & \dir{black}
      & \dir{black} & \con{black} & \dir{black} & 

      & 69\% \\
    FTL & Fortune Telling
      & 
      & \dir{purple} & \inh{purple} & \inh{purple} & \inh{purple} & \inh{purple}
      & \inh{purple} & \inh{purple}

      & 
      &  &  &

      & 
      &  \dir{blue}& \inh{blue} 

      & 
      &  & 

      & 
      & \dir{cyan} & \inh{cyan}

      & 
      &  \dir{black} & \con{black} & \dir{black} & 
      & \dir{black}
      &  & 
      & \dir{black}
      & \dir{black}
      & \dir{black} & \con{black} & \dir{black} & 

      &  69\%     \\
    CAT & Catastrophising
      & 
      &  &  & & & 
      & & \dir{black}

      & 
      &  \dir{red} & \inh{red}  & \inh{red}

      & 
      &  \dir{blue}& \inh{blue} 

      & 
      & \dir{orange} & \inh{orange}

      & 
      & \dir{cyan} & \inh{cyan}

      & 
      &  &  & \dir{black} & \dir{black} 
      & 
      &  & \dir{black}
      & 
      & 
      & \dir{black} & &  & 

      & 48\% \\
    DQP & Disqualifying the Positive
      & 
      &  &  & & & 
      & & 

      & 
      &   & & 

      & 
      &  \dir{blue}& \inh{blue} 

      & 
      &  & 

      & 
      & \dir{cyan} & \inh{cyan}

      & 
      & \dir{black} & \dir{black} & \dir{black} & 
      & \con{black}
      &  & 
      & 
      & \dir{black}
      & \dir{black} & & \dir{black} & \dir{black}

      & 41\% \\
    \midrule
    \multicolumn{38}{@{}l}{\textbf{Occasionally Adopted:} Includes semantically overlapping or synonymous variants, which are often merged in practice.} \\
    \midrule
    MAG & Magnification
      & 
      & \dir{purple} & \inh{purple} & \inh{purple} & \inh{purple} & \inh{purple}
      & \inh{purple} & 

      & 
      &   & & 

      & 
      &  & \dir{black}

      & 
      &  & 

      & 
      &  & \dir{black}

      & 
      & \dir{black} &  & & 
      & 
      &  & 
      & 
      & 
      &  & \con{black} & \dir{black} & 

      & 38\% \\
    JTC & Jumping to Conclusions
      & 
      &  &  &  &  & 
      &  & 

      & 
      & \dir{red}  & \inh{red} & \inh{red}

      & 
      &  & 

      & 
      & \dir{orange} & \inh{orange}

      & 
      &  & 

      & 
      &  & \dir{black} & \dir{black} & 
      & 
      &  & 
      & 
      & 
      &  &  &  & \dir{black}

      & 28\% \\
    BLM & Blaming
      & 
      &  &  &  &  & 
      &  & 

      & 
      &  &  & 

      & 
      & \dir{blue} & \inh{blue}

      & 
      & \dir{orange} & \inh{orange}

      & 
      &  & 

      & 
      &  &  & \dir{black} & \dir{black}
      & 
      &  & 
      & 
      & 
      &  &  & \dir{black} & 

      & 24\% \\
    CMP & Comparing
      & 
      &  &  &  &  & 
      &  & 

      & 
      &  &  & 

      & 
      & \dir{blue} & \inh{blue}

      & 
      &  & 

      & 
      &  & \dir{black}

      & 
      &  &  & \dir{black} & 
      & 
      &  & 
      & 
      & 
      &  &  &  & 

      & 14\% \\
    MXN & Magnification or Minimisation\textsuperscript{}
      & 
      &  &  &  &  & 
      &  & 

      & 
      &  &  & 

      & 
      &  & 

      & 
      &  & 

      & 
      & \dir{cyan} & 

      & 
      &  & \dir{black} &  & 
      & 
      &  & 
      & 
      & \dir{black}
      &  &  &  & \dir{black}

      & 14\% \\
    \midrule
    \multicolumn{38}{@{}l}{\textbf{Rarely Adopted:} Poorly represented in \ac{nlp} studies, often appearing only in isolated datasets.} \\
    \midrule
    BRT & Being Right
      & 
      &  &  &  &  & 
      &  & 

      & 
      &  &  & 

      & 
      &  & 

      & 
      & \dir{orange} & \inh{orange}

      & 
      &  & 

      & 
      &  &  &  & \dir{black}
      & 
      &  & 
      & 
      & 
      &  &  &  & 

      & 10\% \\
    CTL & Control Fallacy
      & 
      &  &  &  &  & 
      &  & 

      & 
      &  &  & 

      & 
      &  & 

      & 
      & \dir{orange} & \inh{orange}

      & 
      &  & 

      & 
      &  &  &  & \dir{black}
      & 
      &  & 
      & 
      & 
      &  &  &  & 

      & 10\% \\
    FOC & Fallacy of Change
      & 
      &  &  &  &  & 
      &  & 

      & 
      &  &  & 

      & 
      &  & 

      & 
      & \dir{orange} & \inh{orange}

      & 
      &  & 

      & 
      &  &  &  & \dir{black}
      & 
      &  & 
      & 
      & 
      &  &  &  & 

      & 10\%     \\
    FOF & Fallacy of Fairness
      & 
      &  &  &  &  & 
      &  & 

      & 
      &  &  & 

      & 
      &  & 

      & 
      & \dir{orange} & \inh{orange}

      & 
      &  & 

      & 
      &  &  &  & \dir{black}
      & 
      &  & 
      & 
      & 
      &  &  &  & 

      & 10\%     \\
  NFE & Negative Feeling or Emotion
      & 
      &  &  &  &  & 
      &  & 

      & 
      &  &  & 

      & 
      & \dir{blue} & \inh{blue}

      & 
      &  & 

      & 
      &  & 

      & 
      &  &  &  & 
      & 
      &  & 
      & 
      & 
      &  &  &  & 

      & 7\% \\
    HRF & Heaven’s Reward Fallacy
      & 
      &  &  &  &  & 
      &  & 

      & 
      &  &  & 

      & 
      &  & 

      & 
      &  & 

      & 
      &  & 

      & 
      &  &  &  & \dir{black}
      & 
      &  & 
      & 
      & 
      &  &  &  & 

      & 3\% \\
    LFT & Low Frustration Tolerance
      & 
      &  &  &  &  & 
      &  & 

      & 
      &  &  & 

      & 
      &  & 

      & 
      &  & 

      & 
      &  & 

      & 
      &  &  & \dir{black} & 
      & 
      &  & 
      & 
      & 
      &  &  &  & 

      & 3\% \\
    MIN & Minimisation
      & 
      &  &  &  &  & 
      &  & 

      & 
      &  &  & 

      & 
      &  & 

      & 
      &  & 

      & 
      &  & \dir{black}

      & 
      &  &  &  & 
      & 
      &  & 
      & 
      & 
      &  &  &  & 

      & 3\% \\
    
    \bottomrule
  \end{tabular}
  \end{adjustbox}
    \caption{
    Consolidated \ac{cd} categories and their inclusion across \ac{nlp} papers in our survey that address the \ac{cd} classification task, either directly or conceptually. Where applicable, papers are grouped by the \ac{cd} dataset they rely on, reflecting the underlying taxonomy of those datasets. Definitions of \ac{cd} categories can be found in Table \ref{tab:cd_definitions}.\\ \textsuperscript{†} Percentage of all papers in this table that reference the given \ac{cd} in any form (\dir{black}, \inh{black}, or \con{black}). \\[0.3em]
    \dir{black} Used in experiments\quad
    \inh{black} Inherited taxonomy from dataset usage\quad
    \con{black} Mentioned conceptually only
    }
  \label{tab:cd_taxonomies}
\end{table*}}

To our knowledge, the most comparable prior survey is by \citet{suputra_2023_survey}, which examined 12 studies and provided an initial synthesis of modelling approaches used at the time. Our work expands on this by covering 38 publications, including recent preprints, and offering a structured overview of the computational landscape for \ac{cd} detection and classification.
This survey aims to provide a clear and practical entry point for researchers engaged in the growing field of \ac{cd} detection and classification from a computational perspective. Our focus is firmly on how computational methods approach these tasks, without seeking to redefine or adjudicate psychological constructs themselves. The contributions in this paper are threefold. 
(i) It provides a consolidated reference of the \ac{cd} taxonomies used across computational studies, highlighting inconsistencies and listing common synonyms (Tables \ref{tab:cd_taxonomies} and \ref{tab:cd_definitions}).
(ii) It defines and analyses task setups, datasets, computational methods, and approaches to performance evaluation, highlighting key patterns and gaps (\S\ref{sec:task-definition}, \S\ref{sec:datasets}, \S\ref{sec:modelling-approaches}, \S\ref{sec:evaluation}).
(iii) It identifies open challenges (\S\ref{sec:challenges_future_directions}) and proposes best practices to guide future research (\S\ref{sec:best-practices}). In doing so, this paper aims to enable more consistent, comparable, and reproducible work in this emerging field.

\section{Task Definitions}
\label{sec:task-definition}

At its core, computational work on \acp{cd} involves a basic classification task: determining whether a given text reflects distorted thinking, and if so, identifying the specific type(s) of distortion. However, a number of different versions of this definition have been adopted in distinct groups of studies, inhibiting direct comparison between them. Some publications combine the classification task with additional tasks that may or may not inform classification. This section reviews how existing research defines \ac{cd} tasks, highlights key differences, and links these to their clinical foundations.

\subsection{Binary Classification (Detection)}

The simplest way to frame the computational task is to ask whether a given text span contains any instance of distorted thinking. This is often called \textit{detection} and is usually treated as a binary classification problem (Distorted vs Undistorted). Methodologically, detection is a basic case of classification with just two labels - the approach remains the same, only the label set is coarser.
This binary framing reflects early clinical aims, where simply noticing the presence of distorted thinking is an important first step before exploring the person’s thoughts in more detail during therapy. Even so, most computational studies go further, aiming to identify specific \ac{cd} types.

\subsection{Single-label vs Multi-label Classification}

While detection addresses the presence of distorted thinking, classification aims to specify which type(s) of \ac{cd} are present in a given text. Here, task definitions diverge based on assumptions around label cardinality:

\paragraph{Single-label classification.} This assumes that each text span reflects exactly one type of distortion. This simplifying assumption is often motivated by practical constraints, such as dataset design or the brevity of inputs (e.g., tweets, short user queries). However, it neglects the fact that distorted thoughts frequently exhibit multiple overlapping \ac{cd} categories, particularly in longer or more detailed texts.

\paragraph{Multi-label classification.} This allows a text to be assigned multiple \ac{cd} categories simultaneously. This formulation more accurately reflects clinical reality, where distortions co-occur and interact. Studies adopting multi-label setups typically model each \ac{cd} type as an independent binary label, which simplifies analysis and inter-annotator agreement calculations \citep{lybarger_2022_identifying, tauscher_2023_automated}.

\subsection{Auxiliary Tasks}

Beyond isolated classification, some studies incorporate auxiliary tasks that extend the utility or interpretability of \ac{cd} models. These are not necessarily distinct problem categories, but rather downstream or complementary tasks that build upon classification outputs:

\paragraph{Reasoning generation.} This involves producing explanatory rationales for why a particular text span was classified as distorted. Methods such as Diagnosis of Thought (DoT) prompting \citep{chen_2023_empowering} and the later ERD (Extraction, Reasoning, Debate) framework \citep{lim_2024_erd} aim to mimic clinician-like reasoning, improving model transparency and trustworthiness.

\paragraph{Reframing generation.} This focuses on producing healthier rephrasings of distorted thoughts, consistent with \ac{cbt} interventions. Studies in this area \citep{sharma_2023_cognitive, maddela-ung_2023_training} treat reframing as a natural extension of detection and classification.

\paragraph{Multi-task learning.} These setups combine \ac{cd} classification with related objectives, such as depression severity prediction \citep{wang_2023_cognitive} or emotion cause extraction \citep{singh_2023_decode}. These formulations tend to leverage the diagnostic value of \ac{cd} features to improve performance on auxiliary tasks.

\paragraph{Multi-modal approaches.} While the vast majority of computational \ac{cd} research focuses on textual data, recent work has begun to explore multi-modal approaches. For example, \citet{singh_2023_decode} integrated text, audio, and video inputs from therapist-patient interactions to enhance \ac{cd} detection. Though still nascent, these efforts highlight the potential of multi-modal signals in capturing the subtle nuances of distorted thinking in real-world settings.

\section{Taxonomies of \acp{cd} in \ac{nlp}}
\label{sec:cd_taxonomies}

Despite the shared objective of classifying \acp{cd}, studies seeking to label \acp{cd} have adopted diverse taxonomies. Without standardisation, it is hard to keep annotations consistent, compare models properly, or clearly interpret the results. 
Table~\ref{tab:cd_taxonomies} illustrates this fragmentation by mapping which \ac{cd} categories are recognised across the papers surveyed. While some categories, such as \textit{Labelling} and \textit{Should Statements}, are commonly used, others are inconsistently applied or redefined in various ways.
Early conceptions of \acp{cd} were proposed by \citet{beck_1963_thinking}, who described patterns of dysfunctional thinking with examples such as arbitrary inference and overgeneralisation. This work was later popularised by \citet{burns_1999_feeling}, whose ten-category taxonomy is frequently cited in psychology literature.\footnote{While \citet{burns_1999_feeling} is not the only taxonomy of \acp{cd} used in psychological contexts, and clinical practice often employs broader or alternative frameworks, our focus is on the inconsistency in how these definitions are interpreted and applied within computational research.} However, computational studies do not uniformly follow this framework.

Some papers, such as \citet{shickel_2020_mental-health-text}, draw on definitions from popular psychology sources (e.g., PsychCentral, Psychology Today), resulting in the inclusion of broader or differently framed categories. Other works make subtle assumptions in how the Burns taxonomy is applied, for example, by splitting \textit{Jumping to Conclusions} into its subcategories \textit{Mind Reading} and \textit{Fortune Telling}, sometimes without explicit rationale.

Terminology also varies across studies. The \ac{cd} category referred to as \textit{All or Nothing Thinking} also frequently appears under alternative labels such as \textit{Black and White Thinking}, \textit{Polarised Thinking}, or \textit{Dichotomous Reasoning}. This terminological variety complicates efforts to harmonise and compare datasets.
Since \citet{rojas-barahona_2018_deep-learning} noted that CDs were `fairly well standardised' in computational research, the field has grown considerably - from a handful of studies to 38 now in this survey - leading to a surge in differing taxonomies.
The aim of Table~\ref{tab:cd_taxonomies} is to offer a consolidated view of this evolving landscape, providing a resource for future \ac{nlp} research. Additionally, we include an appendix table (Table~\ref{tab:cd_definitions}) listing synonyms, definitions, and hierarchical relationships between \ac{cd} types, to support more consistent and transparent classification efforts in computational contexts.

\section{Datasets}
\label{sec:datasets}

{

    
    \begin{table*}[t]
      \centering
      \scriptsize
      \begin{adjustbox}{width=\textwidth}
      \begin{threeparttable}
      \begin{tabular}{@{}l l r l l l@{}}
      \toprule
      \textbf{Dataset}\textsuperscript{†} & \textbf{Language} & \textbf{Size (\# Samples)\textsuperscript{*}} & \textbf{Labelling}\textsuperscript{‡} & \textbf{Annotators} & \textbf{Access} \\
      \midrule
      \multicolumn{6}{@{}l}{\textbf{Literature Examples}} \\
      \midrule
      \citet{wiemer_2004_automatic} & English & 261 & Single-label (10) & Expert & Private \\
      \midrule
      \multicolumn{6}{@{}l}{\textbf{Social Media}} \\
      \midrule
      \citet{alhaj_2022_improving} & Arabic & 9,250 & Single-label (5) & Non-Expert (Unspecified) & Private \\
      \textsc{SocialCD-3K}, \citet{qi_2024_supervised} & Mandarin & 3,407 & Multi-label (12) & Domain-Informed & Public \\
      \citet{aureus_2021_determining} & English & 586 & Binary (2) & Mixed & Private \\
      \citet{simms_2017_ml} & English & 459 & Binary (2) & Mixed & Private \\
      \midrule
      \multicolumn{6}{@{}l}{\textbf{Digital Mental Health Platform}} \\
      \midrule
      \citet{rojas-barahona_2018_deep-learning} & English & 4,035 & Multi-label (15) & Expert & Private \\
      \citet{lin_2024_detection} & Mandarin & 4,001 & Binary (2) & Domain-Informed & Public \\
      \textsc{TherapistQA}, \citet{shreevastava_2021_detecting} & English & 2,529 & Multi-label (10) & Non-Expert (Unspecified) & Public \\
      \textsc{MH-D}, \citet{shickel_2020_mental-health-text} & English & 1,799 & Binary (2) & Domain-Informed & Private \\
      \textsc{MH-C}, \citet{shickel_2020_mental-health-text} & English & 1,164 & Single-label (15) & Domain-Informed & Private \\
      \textsc{CBT-CD}, \citet{zhang_2025_cbt} & English & 146 & Multi-label (10) & Expert & Public \\
      \midrule
      \multicolumn{6}{@{}l}{\textbf{Crowd-sourced}} \\
      \midrule
      \citet{elsharawi_2024_cjournal} & English & 34,370 & Single-label (14) & Expert & Private \\
      \textsc{PatternReframe}, \citet{maddela-ung_2023_training} & English & 9,688 & Multi-label (10) & Crowd-Generated & Public \\
      \textsc{CrowdDist}, \citet{shickel_2020_mental-health-text} & English & 7,666 & Single-label (15) & Crowd-Generated & Private \\
      \textsc{C2D2}, \citet{wang_2023_c2d2} & Mandarin & 7,500 & Single-label (7) & Crowd-Generated & Request \\
      \textsc{Thinking Trap}, \citet{sharma_2023_cognitive} & English & 600 & Multi-label (13) & Expert & Public \\
      \midrule
      \multicolumn{6}{@{}l}{\textbf{Synthetic}} \\
      \midrule
      \textsc{GPT-4 Synthetic}, \citet{babacan_2025_creating} & English & 2,000 & Single-label (10) & Automated (LLM) & Public \\
      \midrule
      \multicolumn{6}{@{}l}{\textbf{Clinical Intervention}} \\
      \midrule
      \citet{lalk_2024_depression} & German & 104,557 & Multi-label (14) & Automated (Lexicon) & Request \\
      \citet{lybarger_2022_identifying} & English & 7,436 & Multi-label (5) & Expert & Private \\
      \midrule
      \multicolumn{6}{@{}l}{\textbf{Hybrid (Mixed Domains)}} \\
      \midrule
      \textsc{KoACD}, \citet{kim_2025_koacd} & Korean & 108,717 & Single-label (10) & Automated (LLM) & Request \\
      \textsc{GPT-4 Combined}, \citet{babacan_2025_creating} & English & 4,530 & Single-label (10) & Automated (LLM) & Request \\
      \textsc{CoDEC}, \citet{singh_2023_decode} & English & 3,773 & Binary (2) & Non-Expert (Unspecified) & Request \\
      \textsc{CoDeR}, \citet{singh_2024_deciphering} & English & 3,773 & Binary (2) & Trained & Public \\
      \citet{wang_2023_cognitive} & English & 3,644 & Single-label (11) & Automated (BERT) & Private \\
      \citet{mostafa_2021_automatic} & English & 2,409 & Single-label (2) & Domain-Informed & Private \\
      \bottomrule
      \end{tabular}

      \caption{
        Overview of datasets for \ac{cd} detection and classification, grouped by domain.  
        See Appendix Table~\ref{tab:cd_datasets_full} for an expanded version with agreement metrics, access details, and subdomains.  
        \textsuperscript{†} Corpus name, or earliest study to use it for \ac{cd} tasks.  
        \textsuperscript{*} Number of annotated units (e.g., posts, speech turns); for automated annotations, items processed.  
        \textsuperscript{‡} Number of \ac{cd} categories used, excluding `Undistorted' for classification.
    }
    
      \label{tab:cd_datasets}
      \end{threeparttable}
      \end{adjustbox}
      
    \end{table*}

}

Datasets are the foundation for research on \ac{cd} classification, providing the labelled examples needed to develop and evaluate detection methods. However, existing datasets vary widely in scope, annotation practices, and accessibility. To organise this diversity, we group datasets by their underlying data sources and contexts of use.

\subsection{Domains}

We use the term \textit{domain} to describe the broader context from which a dataset’s text data originates. Domains shape the linguistic style of examples, affect annotation reliability, and carry practical considerations such as data privacy and availability. The following six domains reflect the main sources of data in current \ac{cd} classification research.

\paragraph{Literature Examples.} Early work, such as \citet{wiemer_2004_automatic}, used \ac{cd} examples from existing psychological literature \citep{beck_1979_cognitive, burns_1999_feeling}. While these examples are clear and well-labelled, they are typically idealised and explicit, limiting their applicability to real-world, patient-generated language.

\paragraph{Social Media Platforms.} Public posts from platforms such as Reddit \citep{aureus_2021_determining}, Twitter \citep{alhaj_2022_improving}, and Weibo \citep{qi_2024_supervised} provide naturally occurring \ac{cd} instances in user-generated content. This domain offers large volumes of data but poses challenges related to linguistic noise and context ambiguity.

\paragraph{Digital Mental Health Platforms.} Peer-support services, such as Koko \citep{rojas-barahona_2018_deep-learning} and TaoConnect \citep{shickel_2020_mental-health-text}, have been valuable sources of data rich in \acp{cd}. The widely used \textsc{TherapistQA} dataset \citep{shreevastava_2021_detecting} originates from a Kaggle Q\&A repository and has since been extended in multiple studies \citep{chen_2023_empowering, babacan_2025_creating, lalk_2024_depression}.

\paragraph{Crowd-Sourced Approaches.} To tackle issues of data scarcity and privacy, several studies have turned to crowdworkers to generate or annotate \ac{cd} examples. Well-known corpora created this way include \textsc{CrowdDist} \citep{shickel_2020_mental-health-text}, \textsc{PatternReframe} \citep{maddela-ung_2023_training}, and \textsc{Thinking Trap} \citep{sharma_2023_cognitive}. These datasets are scalable and flexible but often lack the subtlety of real-world data, as crowdworkers may produce overly explicit examples.

\paragraph{Clinical Interventions.} These datasets, derived from real therapeutic conversations, reflect how people communicate in real-world settings. Notable examples include annotated patient-therapist text message exchanges \citep{lybarger_2022_identifying, tauscher_2023_automated} and psychotherapy transcripts \citep{lalk_2024_depression}. Multimodal corpora such as \textsc{CoDEC} and \textsc{CoDeR} \citep{singh_2023_decode, singh_2024_deciphering} also fall into this category, although they involve a mix of authentic and staged interactions. Despite their value, these datasets are often subject to access restrictions due to privacy and ethical considerations.

\paragraph{Synthetic Datasets} Recent work has explored using \acp{llm} to generate synthetic \ac{cd} data. \citet{babacan_2025_creating} created GPT-4-generated corpora, while \citet{kim_2025_koacd} recently released \textsc{KoACD}, a Korean dataset augmenting social media data with synthetic samples. Synthetic datasets support balanced and scalable resource creation, but may fail to capture the nuanced linguistic and contextual patterns present in genuine human language.

\subsection{Comparative Discussion}

Each domain offers distinct strengths and drawbacks. Clinical intervention datasets are highly representative of real-world therapeutic contexts but are usually small and difficult to obtain. In contrast, social media and digital mental health platforms provide scalable, naturally occurring data, though they often exhibit linguistic noise and structural inconsistency. Crowd-sourced datasets allow for controlled creation of \ac{cd} examples but can introduce stylistic artefacts that may not mirror authentic language use. Synthetic datasets, including those generated by \acp{llm}, support large-scale experimentation and balanced class distributions, yet require thorough validation to ensure their realism. Most existing corpora are monolingual, predominantly in English, as summarised in Table~\ref{tab:cd_datasets}. However, recent efforts have started expanding into other languages. For instance, \citet{wang_2023_c2d2} introduced \textsc{C2D2}, a Mandarin corpus, while \citet{kim_2025_koacd} developed \textsc{KoACD}, a Korean dataset. Additional dataset details, including subdomains, access links, and \ac{iaa} figures, are provided in Appendix Table~\ref{tab:cd_datasets_full}.

\subsection{Annotation Strategies}

Annotation strategies vary considerably across datasets. Clinical and literature-derived corpora typically rely on expert annotators, prioritising label quality at the expense of scalability. In contrast, social media and crowd-sourced datasets often involve non-expert annotators, sometimes supported by brief, domain-specific training from qualified psychologists to improve consistency. While these strategies enable large-scale annotation, reliably classifying \acp{cd} remains challenging. The task demands subtle, often subjective judgements, and studies consistently report low \ac{iaa}, particularly when annotators lack deeper domain expertise. To mitigate this, some works adopt strict inclusion criteria, only retaining examples where annotators fully agree on the label or a subset of distortion types \citep{aureus_2021_determining, shickel_2020_mental-health-text}. Though this approach improves label precision, it risks introducing bias by systematically excluding ambiguous or borderline cases - which are arguably the most reflective of real-world \ac{cd} occurrences.

\section{Modelling Approaches}
\label{sec:modelling-approaches}

This section outlines computational approaches to \ac{cd} detection and classification, grouped into six methodological categories that reflect major developments in the field.

\subsection{Rule-Based}

The first systems for \ac{cd} classification were rule-based, using hand-crafted keyword patterns and syntactic features. \citet{wiemer_2004_automatic} developed COGNO, a system that mapped surface linguistic cues (e.g. verb tense, negation, person markers) to predefined \ac{cd} categories. It performed well on a 10-class single-label task (Macro-F1 = 0.61) but was only tested on `polished' textbook-style \ac{cd} examples. One of the main strengths of rule-based systems is their interpretability, a feature still highly valued in clinical settings, where transparency is critical. For instance, \citet{lalk_2024_depression} employed a manually curated list of n-grams, based on previous work \citep{bathina_2021_depressed}, to monitor distortion frequency in psychotherapy transcripts and predict patient depression severity. 

\subsection{Traditional Machine Learning (Feature-based)}

As more annotated corpora became available, early rule-based systems gave way to feature-based statistical models. These approaches combined classic classifiers, such as logistic regression (LR) and support vector machines (SVMs), with engineered features such as Linguistic Inquiry and Word Count (LIWC) scores and Term Frequency-Inverse Document Frequency (TF-IDF) vectors. \citet{simms_2017_ml} demonstrated that LR trained on LIWC features performed well on the detection task using Tumblr data, while \citet{shickel_2020_mental-health-text} found that TF-IDF with LR outperformed CNNs on a 15-class single-label synthetic dataset (F1 = 0.68), indicating that shallow linguistic cues can remain competitive even in classification tasks. Similarly, \citet{shreevastava_2021_detecting} used SVM with smooth inverse frequency (SIF) embeddings, achieving strong performance (F1 = 0.77) on the detection task despite the insensitivity of SIF to word order.

\subsection{Deep Learning with Static Embeddings}

Static word embeddings such as Word2Vec \citep{mikolov_2013_word2vec} and GloVe \citep{pennington-2014-glove} introduced vector-based representations of words derived from co-occurrence patterns. These effectively captured word similarity but failed to account for contextual nuance (e.g., `riverbank' vs `bank'). Despite this limitation, such embeddings formed the backbone of early deep-learning based \ac{cd} classifiers. For example, \citet{rojas-barahona_2018_deep-learning} combined GloVe with CNNs for multi-label \ac{cd} classification, outperforming traditional models. Similarly, \citet{mostafa_2021_automatic} trained LSTM models using GloVe vectors to classify different \ac{cd} types, although their findings highlighted concerns around overfitting due to the use of limited and synthetic data. To address data sparsity in Arabic texts, \citet{alhaj_2022_improving} applied contextual topic modelling (CTM), integrating static embeddings with domain-specific topic information via BERTopic - an approach that proved helpful in low-resource settings. While static embeddings offered richer representations than earlier methods, their inability to capture context, especially for polysemous words and subtle pragmatic distinctions, ultimately led to the rise of contextual models.

\subsection{Transformer-based Architectures}

Transformers, particularly BERT (Bidirectional Encoder Representations from Transformers) \citep{devlin_2019_bert}, introduced contextual embeddings that capture word meaning based on the surrounding sentence, marking a significant leap forward in many \ac{nlp} tasks, including \ac{cd} detection and classification. \citet{shreevastava_2021_detecting} used fine-tuned Sentence-BERT (SBERT) for binary \ac{cd} detection, showing notable improvements over earlier models. Similarly, \citet{lybarger_2022_identifying} demonstrated that incorporating conversational history further improves classification performance on therapy dialogues. Domain-adapted transformers like MentalBERT \citep{ji_2022_mentalbert}, which are pre-trained on mental-health-related data, showed benefits over general-purpose models, while augmentation techniques such as mixup \citep{zhang_2018_mixup} were explored to improve performance on rare \ac{cd} classes (minor improvements). \citet{maddela-ung_2023_training} found that models like RoBERTa performed better than larger language models such as GPT-3.5 for this task, though distinguishing between closely related distortions remained a challenge. Overall, transformers brought significant gains in robustness and accuracy, but these came at the cost of reduced interpretability and a higher risk of overfitting, especially in datasets with extreme class imbalance. These limitations set the stage for the emergence of prompt-based models.

\subsection{LLMs and Prompting Frameworks}

\acp{llm} like GPT-3 have enabled \ac{cd} detection through natural language prompting, allowing models to perform the task without the need for tailored training. Recent work in this area can broadly be divided into two approaches: zero-shot prompting and chain-of-thought frameworks.

\paragraph{Zero-shot prompting.} \citet{chen_2023_empowering} introduced Diagnosis-of-Thought (DoT) prompting, guiding \acp{llm} to reason through \ac{cd} detection and classification with structured outputs. While this method sometimes outperformed fine-tuned transformers on the \textsc{TherapistQA} corpus, it also is prone to hallucinations and inconsistent rationales, especially on ambiguous cases. Similarly, \citet{pico_2025_comparative} compared multiple \acp{llm}, finding that well-prompted open-source models could approach the performance of larger proprietary models, though results were not always consistent across runs.

\paragraph{Chain-of-thought (CoT).} \citet{lim_2024_erd} presented the ERD framework, where multiple \ac{llm} agents simulate therapist-like reasoning, extracting emotional cues and providing structured explanations. Though this method produced richer rationales, it was highly sensitive to how prompts were designed and faced challenges in scalability and validation. Another key application of \acp{llm} has been synthetic data generation. For example, \citet{babacan_2025_creating} used GPT-4 to create a balanced \ac{cd} dataset. While initial results were promising, the synthetic data struggled to generalise to the noisier, more complex language found in real-world user inputs.

Overall, prompted \acp{llm} offer advantages in reducing training costs and improving interpretability, but they remain limited by issues such as prompt fragility, hallucinations, and inconsistent evaluation results.

\subsection{Multimodal and Multi-task Architectures}

To overcome the limitations of text-only models, recent research has explored incorporating additional modalities and joint tasks. \citet{singh_2023_decode} introduced \textsc{CoDEC}, a multimodal dataset combining video, audio, and text from therapy simulations. By leveraging intonation and facial expressions, their model achieved improvements in detecting emotion-related \acp{cd} such as \textit{Emotional Reasoning}. Building on \textsc{CoDEC}, \citet{singh_2024_deciphering} developed \textsc{CoDeR}, which added annotated reasoning spans to support explanation-aware \ac{cd} classification. Multi-task learning has also been employed to utilise the diagnostic value of \acp{cd}. For example, \citet{lee_2021_micromodels} repurposed micromodel outputs to improve depression and PTSD prediction, while \citet{wang_2023_c2d2} demonstrated that incorporating \ac{cd} frequency improved mental illness detection pipelines. These architectures have shown promise in boosting robustness, particularly in low-resource or noisy data settings. However, the scarcity of multimodal datasets and challenges around annotation and privacy continue to hinder wider adoption.

\subsection{Feasibility of Meta-Analysis}

While we initially intended to include comparative performance tables, meaningful aggregation proved infeasible. The studies diverge across critical axes - (i) task formulation, (ii) \ac{cd} taxonomy choice, (iii) dataset domain, (iv) evaluation metrics, (v) granularity of the unit of analysis, (vi) context inclusion and window size, and (vii) modality - often incompatibly. Sample sizes within aligned subgroups are too small for robust comparison, and \ac{iaa} is reported inconsistently, further limiting comparability. Pooled tables would therefore risk suggesting misleading trends. We release the full set of extracted results in our GitHub repository, enabling researchers to build their own comparisons. In this paper, we restrict ourselves to qualitative synthesis, deferring formal meta-analysis until the evidence base is larger and more standardised.

\section{Evaluation}
\label{sec:evaluation}

Despite progress in \ac{cd} classification, evaluation practices remain inconsistent, with studies differing markedly in their choice and reporting of metrics, which hinders comparability. Although F1 score is the most commonly used metric, distinctions between macro, micro, and weighted variants are frequently overlooked - a significant issue for class-imbalanced datasets where per-class performance is critical. Some studies now report AUPRC (Area Under the Precision-Recall Curve) to account for skewed label distributions, providing a more informative measure of performance on rare \ac{cd} types \citep{ding_2022_improving}. Nonetheless, per-class metrics are still underreported, hiding weaknesses in addressing infrequent distortions. Similarly, dataset quality is often inconsistently assessed. \Ac{iaa} is either inconsistently measured or reported using incomparable metrics. While Cohen’s Kappa ($\kappa$), which adjusts for chance, is typically more appropriate, many studies instead rely on raw agreement or non-standard metrics, blurring the line between annotation reliability and model performance. Baseline comparisons are further complicated by inconsistent \ac{cd} taxonomies and datasets.

\section{Challenges \& Future Directions}
\label{sec:challenges_future_directions}

Despite recent progress, the automatic detection and classification of \acp{cd} remains a challenging task, both conceptually and computationally. In this section, we outline three key challenges currently limiting the field: (1) inconsistency in \ac{cd} taxonomies, (2) data scarcity and imbalance, and (3) the overreliance on short-form text. Addressing these issues is essential for improving model performance, evaluation fairness, and eventual clinical applicability.

\subsection{Inconsistent \ac{cd} Taxonomies}

A longstanding challenge is the lack of a standardised taxonomy for \acp{cd}. While foundational frameworks such as the Burns ten-category list \citep{burns_1999_feeling} are commonly cited, computational studies diverge significantly in how they define, split, or rename distortion types. For instance, \textit{Jumping to Conclusions} is frequently subdivided into \textit{Mind Reading} and \textit{Fortune Telling}, and terms such as \textit{All or Nothing Thinking} appear under multiple aliases (e.g., \textit{Black and White Thinking}, \textit{Polarised Thinking}), making it difficult to compare models across studies, reproduce results, or interpret outputs reliably. These inconsistencies also affect annotation quality, as ambiguous or overly granular label sets introduce subjectivity and reduce \ac{iaa} - an issue compounded by the lack of formal guidance on taxonomy use.

\subsection{Data Scarcity, Imbalance, and Annotation Limits}

The field remains constrained by a lack of large, high-quality datasets that capture authentic, context-rich examples of distorted thinking. Clinical corpora are scarce and often inaccessible due to privacy constraints, while many widely used datasets are synthetic, crowd-sourced, or compiled from multiple domains, which may lack the nuance and ambiguity of real-world language. This limits the complexity of distortions that models can learn and typically results in heavy class imbalance, with rare distortion types being underrepresented or excluded altogether. Although augmentation strategies such as mixup or back-translation offer minor gains for rare categories \citep{ding_2022_improving}, a deeper issue lies in the ceiling imposed by annotation reliability itself. In some settings, even expert annotators show limited agreement, particularly for subtle or overlapping categories. For instance, \citet{tauscher_2023_automated} report that for the presence of `Any Distortion' in a text, human F1 agreement was 0.63, while a fine-tuned BERT model reached 0.62 - a very small difference. This suggests that for certain formulations, such as binary detection or high-frequency classes, current models may already be approaching the upper bound set by annotation quality. It also reinforces the need for clearer task definitions and more consistent annotation protocols before investing in model complexity.

\subsection{Overreliance on Short Text}

The vast majority of existing datasets frame \ac{cd} detection at the sentence or single-utterance level. This simplifies annotation and model design but introduces strong limitations, as many distortions are context-dependent or only weakly signalled lexically. By stripping away discourse-level cues, models are forced to rely on surface-level patterns and may perform poorly on more ambiguous cases. Predictably, this has contributed to the dominance of distortion types with overt markers (e.g., \textit{Should Statements}) in both datasets and model outputs. Empirical studies confirm the value of richer context, showing that including prior conversational turns improved detection F1 from 0.68 to 0.73 \citep{lybarger_2022_identifying}, while frameworks such as ERD achieve greater interpretability by explicitly reasoning over multi-sentence inputs \citep{lim_2024_erd}. Still, most benchmarks continue to prioritise short-form inputs. Moving forward, we argue that context-aware models should become the norm rather than the exception. In parallel, new datasets should prioritise multi-turn conversations, real patient narratives, and longer-form content that more closely mirrors therapeutic language.

\section{Best Practices and Recommendations} \label{sec:best-practices}

Alongside consolidating existing research, it is important to address the main sources of fragmentation in the field. We therefore propose a set of best practices for future work, which should be followed where possible or clearly justified if deviated from.

\subsection{Taxonomy Adoption}

As shown in Table~\ref{tab:cd_taxonomies}, inconsistent use of \ac{cd} taxonomies has made cross-study comparison difficult. In the absence of a universally accepted taxonomy, we recommend Burns’ taxonomy\footnote{We provide recommended label sets online.} \citep{burns_1999_feeling} as a sensible default, since it is the most widely cited and most alternatives used in \ac{nlp} are partial reinterpretations of it. Researchers should (i) report the source and rationale for their chosen taxonomy, (ii) avoid introducing new or expanded taxonomies without justification, and (iii) prioritise taxonomies grounded in clinical consensus. Reliance on loosely defined online taxonomies is discouraged. Where deviations from Burns' taxonomy are necessary, the rationale should be documented in the study, or, in the case of new datasets, in the corresponding datasheet, for which we provide a template online.

\subsection{Unambiguous Evaluation Reporting}

Inconsistent reporting is a major barrier to comparing results across studies. To improve comparability, we recommend that future work (i) clearly state the task formulation (detection, single-label, or multi-label classification; \S\ref{sec:task-definition}), (ii) specify the analysis unit (sentence, turn, session) and, if relevant, document the exact context window, (iii) indicate the \ac{cd} taxonomy used, (iv) report per-class scores alongside macro and weighted F1\footnote{For imbalanced or multi-label settings, AUPRC (micro or macro) and per-class PR curves may be more informative. Ultimately, metrics should fit the use case.} with unambiguous labels, and (v) explain the choice of evaluation metrics. While a single metric cannot suit all applications, departures from macro or weighted F1 should always be accompanied by a clear rationale.

\subsection{Dataset Development and Use}

For researchers creating new \ac{cd} datasets, we recommend providing a datasheet that documents the dataset’s origin, annotation protocol, size, taxonomy, analysis unit, and licensing. To support this, we provide a standardised datasheet template in our GitHub repository. For reuse of existing datasets, we encourage researchers to apply our \textit{Ethics Flowchart}, which provides practical guidance on assessing provenance, consent, and documentation before experimentation.

\subsection{Annotation Reliability and Inter-annotator Agreement}

Annotation processes should be reported transparently. Researchers should provide standard \ac{iaa} metrics (e.g., Cohen’s $\kappa$, Fleiss’ $\kappa$, Krippendorff’s $\alpha$) rather than vague statements of `agreement'. The rationale for the chosen metric should be stated, and partial dataset sampling for \ac{iaa} is acceptable provided that procedures are clearly documented.

We further recommend that future work (i) reports human-model performance comparison metrics where possible, (ii) prioritises the release of multilingual and multi-domain corpora, and (iii) explicitly documents how annotation disagreements are resolved. Without such practices, gains in model performance may reflect noise-fitting rather than genuine progress.

\subsection{Code and Dataset Release}

To support replication and benchmarking, researchers should release code and, where licensing and privacy considerations permit, datasets. As shown in Table~\ref{tab:cd_datasets_full}, many of the surveyed studies do not provide public implementations. We recommend that future work adopt code and data release as standard practice, in line with recent broader calls for stronger reproducibility standards in AI research and governance \citep{semmelrock_2025_reproducibilitymachinelearningbasedresearch, mason-williams_2025_reproducibility}.


\section{Limitations}

While this survey offers a structured overview of methods, datasets, and evaluation practices for \ac{cd} detection and classification in \ac{nlp}, it has several limitations. The focus is primarily computational, with limited integration of insights from clinical psychology or cognitive science, and deeper conceptual analyses of \acp{cd} are beyond its scope. The survey also centres on English-language datasets and approaches, which may limit generalisability to other languages and cultural contexts. Although emerging work on multimodal and conversational systems is noted, the emphasis remains on text-based methods and classification tasks, rather than auxiliary tasks such as cognitive reframing. Finally, due to space constraints, some datasets and methods could not be covered in detail, and despite systematic efforts, some relevant studies may have been missed.

As with any literature survey, our analysis is constrained by the scope, reporting quality, and coverage of the included studies, and should be viewed as a snapshot of a rapidly evolving field. As such, some very recent preprints may not be included. While we have aimed for balanced representation, our synthesis reflects our methodological choices and interpretive framing, which may influence the emphasis placed on particular themes.

\section{Ethical Considerations}
\label{sec:ethics}

Given the nature of \acp{cd} within psychotherapy contexts, this survey acknowledges several important ethical considerations. As our work is a synthesis of existing studies, we did not collect new data or propose new models. Nevertheless, the scope of our review touches on certain areas of concern that warrant attention.

\paragraph{Dataset Origins.}

Many of the datasets discussed in this survey are derived from sources where individuals may have disclosed personal, and often highly sensitive, information. This is particularly true in domains such as digital mental health platforms, social media, and therapy transcripts. In reviewing these studies, we noted that some datasets have limited publicly available information on certain aspects of their origins or collection processes, often due to constraints inherited from upstream sources.

For instance, with \textsc{TherapistQA}, the authors provide clear documentation - including detailed labelling guidelines and procedures for resolving disagreements between annotators \citep{shreevastava_2021_detecting}. However, because the dataset draws on an upstream public source, some provenance details reflect the level of information made available by that original source rather than any omission by the curators themselves. In this case, the publicly available version is based on a \textit{Kaggle Q\&A} dataset,\footnote{\url{https://www.kaggle.com/datasets/arnmaud/therapist-qa/data}} for which we have not found publicly accessible details specifying the original platform or data collection process. This situation is not unique to \textsc{TherapistQA}; several widely used mental health corpora draw on similar repositories, highlighting a broader and ongoing challenge in achieving complete transparency of data origins within the field.

\paragraph{Linguistic \& Cultural Biases.}

The literature we surveyed remains heavily focused on English-language data, with only limited, though encouragingly increasing, attention paid to other languages or cultural contexts. This linguistic bias introduces significant limitations, especially given that \acp{cd} are almost certainly shaped by cultural norms, stigma, and may manifest quite differently across populations. We repeat calls from prior work for the development and evaluation of more \ac{cd} classification methods that are sensitive to cross-linguistic and cross-cultural variation.

\paragraph{Risks of Misuse \& Overreliance.}

We also acknowledge that the automatic detection of \acp{cd} carries serious risks if applied irresponsibly, particularly outside of therapeutic settings. Misclassification or over-reliance on automated outputs could result in harm - reinforcing stigma, invalidating personal experiences, or leading to inappropriate interventions. We therefore stress that \ac{cd} classification systems should not be deployed without careful validation, the involvement of mental health professionals, and appropriate safeguards to protect user autonomy and well-being. As highlighted in Section \ref{sec:modelling-approaches}, \ac{cd} classification performance is variable and often limited on rarer classes. As such, applications in clinical settings should be approached cautiously.

In presenting this survey, our aim is to support the \ac{nlp} community in enabling more ethically considerate, transparent, and responsible research practices - particularly when working in sensitive domains such as mental health.

\section*{Acknowledgments}

This work was supported by the Engineering and Physical Sciences Research Council [grant number EP/W524475/1]. We thank the anonymous reviewers for their constructive feedback.

\bibliography{custom}

\appendix

\section{Survey Methodology}
\label{sec:survey-methodology}

To compile a comprehensive list of relevant research publications, we drew from the following sources:

\begin{enumerate}
\item Searches conducted across the ACL Anthology\footnote{\url{https://aclanthology.org/}}, arXiv\footnote{\url{https://arxiv.org/}}, PubMed\footnote{\url{https://pubmed.ncbi.nlm.nih.gov/}}, and IEEE Xplore\footnote{\url{https://ieeexplore.ieee.org/}}, with no date restrictions. Search queries included terms such as \textit{`cognitive distortion'} and \textit{`dysfunctional thought'}.
\item Additional papers identified organically via Google Scholar, Semantic Scholar, and reference lists from relevant work.
\end{enumerate}

After manual filtering, we retained 38 primary publications and preprints, spanning from 2004 to May 2025. Studies were included if they (i) implemented a \ac{cd} detection or classification model, (ii) introduced a \ac{cd}-related dataset, or (iii) computationally explored the taxonomy of \acp{cd}. To support reproducibility, we release our supplementary resources and paper list, along with corrections and updates, on GitHub: \url{https://github.com/archiesage/cognitive-distortion-nlp-survey}.

\section{Psychological Foundations of \acp{cd}}
\label{sec:psychology-foundation-appendix}

This appendix offers an overview of \acp{cd}, providing additional context for \ac{nlp} researchers who may be unfamiliar with them.

\subsection{What are \acp{cd}?}

As introduced in Section~\ref{sec:introduction}, \acp{cd}, sometimes called \textit{thinking errors}, are habitual patterns of negatively biased or flawed thinking that shape how people interpret events, evaluate themselves, and respond to the world \citep{beck_1963_thinking}. There are many different types of \acp{cd}, as outlined in Table~\ref{tab:cd_definitions}. In this paper, we use the term \textit{\ac{cd} taxonomy} to refer to the particular set of \acp{cd} adopted in any given \ac{nlp} study.

\acp{cd} are not restricted to clinical settings, as most people exhibit distorted thinking, often automatically, in response to certain situations (a key aspect studied in \ac{cbt}). However, in conditions such as depression, anxiety, and post-traumatic stress disorder, these patterns tend to occur more frequently \citep{lalk_2024_depression}, become harder to shift, and carry a heightened emotional impact.

\subsection{Origins}

The origins of \acp{cd} are often traced back to the early work of Beck in the 1960s. While he identified fewer types of distortions than are commonly recognised today, he did describe well-known types such as \textit{Overgeneralisation}, as well as others that are less frequently cited. Quoted directly from his work \citep{beck_1963_thinking}:\footnote{Both \textit{Arbitrary Interpretation} and \textit{Selective Abstraction} align reasonably well with the consolidated taxonomy reference in Table~\ref{tab:cd_taxonomies}, corresponding to \textit{Jumping to Conclusions} and \textit{Mental Filter}, respectively.}

\begin{itemize}
    \item \textit{Arbitrary Interpretation} - ‘the process of forming an interpretation of a situation, event, or experience when there is no factual evidence to support the conclusion, or when the conclusion is contrary to the evidence.’
    \item \textit{Selective Abstraction} - ‘focusing on a detail taken out of context, ignoring other more salient features of the situation, and conceptualising the whole experience on the basis of this element.’
\end{itemize}

The concept of \acp{cd} gained further traction in the 1980s through the work of David Burns, who outlined a widely used taxonomy of ten \acp{cd} in \textit{Feeling Good: The New Mood Therapy}, later revising it in 1999 \citep{burns_1999_feeling}. Burns’ list more closely resembles the taxonomies commonly used in \ac{nlp} studies today.

\subsection{Context \& Clinical Relevance}

Although this survey focuses on computational approaches to \ac{cd} detection and classification, it is important to situate this area within its broader psychological context. Several therapeutic frameworks address distorted thinking, either directly or indirectly. For example:

\paragraph{Rational Emotive Behaviour Therapy (REBT)} where the focus is on the identification and challenge of irrational beliefs, which are broader than the discrete thought patterns typically considered as \acp{cd}. REBT focuses on core value beliefs (e.g., `I must be the best') that tend to underlie many distortions, aiming to replace them with rational alternatives \citep{ellis_1957_rational, ellis_1994_reason, dryden_2021_rational}.

\paragraph{Schema therapy} moves past the present-moment focus of traditional \ac{cbt} by addressing deep rooted, and often unhelpful, patterns, which are known as schemas. These can develop in childhood when core emotional needs are not met. Such schemas can lead to ongoing problems in how a person thinks, feels, behaves, and relates to others, often requiring longer term and more intensive treatment \citep{young_2006_schema}.

\paragraph{Acceptance and Commitment Therapy (ACT)} does not frame problematic thinking in terms of \acp{cd}, but instead chooses to recognise unhelpful thoughts as a normal part of human thinking. The focus is not on \textit{challenging} the content of these thoughts, but on changing the individual’s relationship to them through various strategies \citep{hayes_2011_acceptance}.

\paragraph{}While these approaches differ in focus, they generally agree on the importance of recognising distorted thinking patterns as a route to improved emotional or behavioural regulation. This clinical grounding remains a key motivation behind the computational modelling of \acp{cd}.

\section{Additional Tables}
\label{sec:cds}

{
    \renewcommand{\arraystretch}{1.1}

    \begin{table*}[t]
        \centering
        \scriptsize
        \begin{adjustbox}{width=\textwidth}
        \begin{tabular}{l l p{3.5cm} p{4.5cm} p{3.5cm}}
        \toprule
        \textbf{Code} & \textbf{Cognitive Distortion} & \textbf{Description} & \textbf{Example} & \textbf{Synonyms} \\
        \midrule
        \multicolumn{5}{l}{\textbf{Burns’ Taxonomy Distortions \citep{burns_1999_feeling}}} \\
        \midrule
        AON & All or Nothing Thinking & Viewing situations in black-and-white terms, without acknowledging nuance or grey areas. & \textit{Since our method didn’t outperform all baselines in every metric, the entire study feels like a failure.} & Black and White Thinking, Polarised Thinking, Dichotomous Reasoning \\
        DQP & Disqualifying the Positive & Rejecting positive outcomes or feedback as unimportant, accidental, or unearned. & \textit{Our paper was accepted, but probably only because the reviewers didn’t scrutinise it deeply enough.} & Discounting the Positive \\
        EMR & Emotional Reasoning & Believing that negative emotions reflect objective truths. & \textit{I feel uneasy about presenting this model, so it must be inherently flawed in ways I’m not seeing.} & \\
        FTL & Fortune Telling\textsuperscript{†} & Predicting negative outcomes as inevitable, without sufficient evidence. & \textit{Given how niche our contribution is, there’s no chance it will get noticed by the review committee.} & Negative Predictions, The Fortune Teller Error \\
        JTC & Jumping to Conclusions\textsuperscript{†} & Making assumptions with insufficient evidence. & \textit{The editor's brief reply likely means they’ve already decided to reject our manuscript.} & Jumping to Negative Conclusions \\
        LBL & Labelling & Defining oneself or others by a single trait or outcome. & \textit{I misinterpreted that reviewer comment, clearly I’m not cut out for academic writing.} & Global Labelling, Labelling and Mislabelling \\
        MAG & Magnification\textsuperscript{*} & Exaggerating the significance of errors or flaws. & \textit{This small formatting mistake will probably make the reviewers think we lack attention to detail.} & Catastrophising\textsuperscript{*} \\
        MIN & Minimisation & Downplaying the significance of positive outcomes, achievements, or strengths, reducing their perceived value or relevance. & \textit{Sure, the paper was accepted, but it didn't get the best reviews, so it doesn’t really count as a proper success.} & \\
        MTF & Mental Filter & Focusing exclusively on negative details. & \textit{One weakness in our ablation study keeps bothering me, despite the overall positive experimental results.} & Filtering \\
        MDR & Mind Reading\textsuperscript{†} & Assuming you know what others are thinking, often negatively. & \textit{The session chair looked disinterested, our work must have been irrelevant to the audience.} & \\
        OVG & Overgeneralisation & Drawing broad conclusions from a single incident. & \textit{Since our last submission was desk-rejected, it’s obvious our current work will face the same fate.} & Overgeneralising \\
        PRS & Personalisation & Attributing external events or failures entirely to oneself. & \textit{The collaboration didn’t materialise, probably because my proposal wasn’t convincing enough.} & Personalisation and Blame, Personalising, Blaming Oneself \\
        SHD & Should Statements & Holding rigid expectations about how oneself or others ought to behave. & \textit{I should always produce novel ideas quickly, taking this long feels like professional incompetence.} & Shoulds, Inflexibility \\
        \midrule
        \multicolumn{5}{l}{\textbf{Other Distortions}} \\
        \midrule
        BRT & Being Right & Placing too high value on proving yourself correct, often at your own or others' expense. & \textit{I’m certain my annotation guidelines are the best. Any disagreement from the team simply indicates they don’t understand the task properly.} & Always Being Right \\
        BLM & Blaming & Attributing too high responsibility for negative outcomes to others, avoiding self-reflection or your own shared responsibility. & \textit{The demo crashed because the organisers didn’t provide adequate technical support, not because of any oversight on our side.} & Blaming Others \\
        CAT & Catastrophising\textsuperscript{*} & Imagining worst-case scenarios and exaggerating potential negative consequences far beyond their realistic likelihood. & \textit{If this preprint has a minor oversight, it could irreparably damage our lab’s reputation and future collaborations.} & \\
        CMP & Comparing & Measuring self-worth against others in a way that undermines your own accomplishments. & \textit{Another lab published a similar paper first - clearly they’re much more capable researchers than we are.} & Comparing and Despairing, Comparison \\
        CTL & Control Fallacy & Believing either complete control over everything or total helplessness in a situation, without middle ground. & \textit{If I don't oversee every single preprocessing step myself, the entire pipeline will end up flawed.} & \\
        FOC & Fallacy of Change & Assuming others should or will change to meet your own personal expectations. & \textit{If only the dataset creators had annotated according to our taxonomy, our analysis would be so much clearer.} & Control of Fallacies \\
        FOF & Fallacy of Fairness & Presuming life or systems must work in a way that aligns with personal standards of fairness. & \textit{It’s unfair that methodologically weaker papers receive more attention just because they’re trendy.} & \\
        HRF & Heaven’s Reward Fallacy & Expecting a guaranteed reward for one’s hard work. & \textit{After months of hyperparameter tuning, this model surely deserves to be the new state-of-the-art.} & \\
        LFT & Low Frustration Tolerance\textsuperscript{*} & Overestimating the severity of minor inconveniences. & \textit{Dealing with this reviewer rebuttal feels impossible. I can’t imagine going through it again.} & \\
        NFE & Negative Feeling or Emotion & Taking emotional discomfort as proof something is wrong. & \textit{Feeling stuck while writing this paper draft surely means the research itself is inherently flawed.} & \\
        \bottomrule
        \end{tabular}
        \end{adjustbox}
        \captionsetup{font=footnotesize,justification=raggedright,singlelinecheck=false}
        \caption{Categories of \acp{cd} observed in computational research. Descriptions and examples are reflective of common interpretations of these distortions in \ac{nlp} contexts. All examples are fictional and not about any specific work or group. To ensure consistency across studies, we also include synonyms and related terms where applicable. \
        \textsuperscript{†} Jumping to Conclusions (JTC) is frequently considered a parent category that includes Fortune Telling (FTL) and Mind Reading (MDR). \
        \textsuperscript{*} Although Magnification (MAG) and Catastrophising (CAT) are often treated as equivalent, we list them separately to highlight subtle conceptual distinctions, following prior work \citep{lalk_2024_depression, agarwal_2025_exploratory}. Similarly, Low Frustration Tolerance (LFT), while similar to CAT, is presented as a distinct category.}
        \label{tab:cd_definitions}
    \end{table*}
}

{
    \renewcommand{\arraystretch}{1.1}



    \begin{table*}[t]
      \centering
      \scriptsize
      \begin{adjustbox}{width=\textwidth}
      \begin{threeparttable}
      \begin{tabular}{@{}l l p{4cm} r l l l l@{}}
      \toprule
      \textbf{Dataset}\textsuperscript{†} & \textbf{Language} & \textbf{Subdomain} & \textbf{Size (\# Samples)\textsuperscript{*}} & \textbf{Labelling}\textsuperscript{‡} & \textbf{Annotators} & \textbf{Agreement} & \textbf{Access} \\
      \midrule
      \multicolumn{8}{@{}l}{\textbf{Literature Examples}} \\
      \midrule
      \citet{wiemer_2004_automatic} & English & Psychology literature & 261 & Single-label (10) & Expert & -- & Private \\
      \midrule
      \multicolumn{8}{@{}l}{\textbf{Social Media}} \\
      \midrule
      \citet{alhaj_2022_improving} & Arabic & Twitter & 9,250 & Single-label (5) & Non-Expert (Unspecified) & $\kappa=0.817_c$ & Private \\
      \textsc{SocialCD-3K}, \citet{qi_2024_supervised} & Mandarin & Weibo `Zoufan' blog & 3,407 & Multi-label (12) & Domain-Informed & -- & Public\tnote{1} \\
      \citet{aureus_2021_determining} & English & Reddit: r/COVID19\_support & 586 & Binary (2) & Mixed & -- & Private \\
      \citet{simms_2017_ml} & English & Tumblr & 459 & Binary (2) & Mixed & -- & Private \\
      \midrule
      \multicolumn{8}{@{}l}{\textbf{Digital Mental Health Platform}} \\
      \midrule
      \citet{rojas-barahona_2018_deep-learning} & English & Koko & 4,035 & Multi-label (15) & Expert & $\kappa=0.61_c$ & Private\tnote{2} \\
      \citet{lin_2024_detection} & Mandarin & PsyQA counselling forums & 4,001 & Binary (2) & Domain-Informed & $\mathrm{JP}=0.88_d$ & Public\tnote{3} \\
      \textsc{TherapistQA}, \citet{shreevastava_2021_detecting} & English & -- & 2,529 & Multi-label (10) & Non-Expert (Unspecified) & $\mathrm{JP}=0.34_c, 0.61_d$ & Public\tnote{4} \\
      \textsc{MH-D}, \citet{shickel_2020_mental-health-text} & English & TaoConnect & 1,799 & Binary (2) & Domain-Informed & -- & Private \\
      \textsc{MH-C}, \citet{shickel_2020_mental-health-text} & English & TaoConnect & 1,164 & Single-label (15) & Domain-Informed & -- & Private \\
      \textsc{CBT-CD}, \citet{zhang_2025_cbt} & English & Patient-therapist QA & 146 & Multi-label (10) & Expert & -- & Public\tnote{5} \\
      \midrule
      \multicolumn{8}{@{}l}{\textbf{Crowd-sourced}} \\
      \midrule
      \citet{elsharawi_2024_cjournal} & English & -- & 34,370 & Single-label (14) & Expert & -- & Private \\
      \textsc{PatternReframe}, \citet{maddela-ung_2023_training} & English & MTurk, Mephisto & 9,688 & Multi-label (10) & Crowd-Generated & $\alpha=0.355_c$ & Public\tnote{6} \\
      \textsc{CrowdDist}, \citet{shickel_2020_mental-health-text} & English & MTurk & 7,666 & Single-label (15) & Crowd-Generated & -- & Private \\
      \textsc{C2D2}, \citet{wang_2023_c2d2} & Mandarin & -- & 7,500 & Single-label (7) & Crowd-Generated & $\kappa=0.67_c$ & Request\tnote{7} \\
      \textsc{Thinking Trap}, \citet{sharma_2023_cognitive} & English & -- & 600 & Multi-label (13) & Expert & -- & Public\tnote{8} \\
      \midrule
      \multicolumn{8}{@{}l}{\textbf{Synthetic}} \\
      \midrule
      \textsc{GPT-4 Synthetic}, \citet{babacan_2025_creating} & English & GPT-4 & 2,000 & Single-label (10) & Automated (LLM) & -- & Public\tnote{9} \\
      \midrule
      \multicolumn{8}{@{}l}{\textbf{Clinical Intervention}} \\
      \midrule
      \citet{lalk_2024_depression} & German & CBT psychotherapy transcripts & 104,557 & Multi-label (14) & Automated (Lexicon) & -- & Request\tnote{10} \\
      \citet{lybarger_2022_identifying} & English & Patient-therapist text exchanges & 7,436 & Multi-label (5) & Expert & $\kappa=0.53_d$ & Private \\
      \midrule
      \multicolumn{8}{@{}l}{\textbf{Hybrid (Mixed Domains)}} \\
      \midrule
      \textsc{KoACD}, \citet{kim_2025_koacd} & Korean & NAVER Knowledge iN + LLM & 108,717 & Single-label (10) & Automated (LLM) & $\kappa=0.78$ & Request\tnote{11} \\
      \textsc{GPT-4 Combined}, \citet{babacan_2025_creating} & English & GPT-4 synthetic + TherapistQA & 4,530 & Single-label (10) & Automated (LLM) & -- & Request\tnote{12} \\
      \textsc{CoDEC}, \citet{singh_2023_decode} & English & Real + staged patient-therapist videos & 3,773 & Binary (2) & Non-Expert (Unspecified) & $F=0.83_d$ & Request\tnote{13} \\
      \textsc{CoDeR}, \citet{singh_2024_deciphering} & English & Real + staged patient-therapist videos & 3,773 & Binary (2) & Trained & $F=0.83_d$ & Public\tnote{14} \\
      \citet{wang_2023_cognitive} & English & Lit.\ examples + social media augment & 3,644 & Single-label (11) & Automated (BERT) & -- & Private \\
      \citet{mostafa_2021_automatic} & English & Twitter, Surveys, HappyDB & 2,409 & Single-label (2) & Domain-Informed & -- & Private \\
      \bottomrule
    \end{tabular}

        \caption{
            Extended overview of datasets for \ac{cd} detection and classification, grouped by domain.  
            Agreement metrics: $\kappa$ = Cohen’s kappa; $\alpha$ = Krippendorff’s alpha; $F$ = Fleiss’s kappa; JP = joint probability; $_d$ = detection; $_c$ = classification. `--' indicates not applicable or not reported.  
            \textsuperscript{†} Corpus name, or earliest study to use it for \ac{cd} tasks.  
            \textsuperscript{*} Number of annotated units (e.g., posts, speech turns); for automated methods, items processed.  
            \textsuperscript{‡} Number of \ac{cd} categories used, excluding `Undistorted' for classification.
        }

        \vspace{0.2in}
        \begin{tablenotes}[flushleft]
          \footnotesize
          \item[1]  \url{https://github.com/HongzhiQ/SupervisedVsLLM-EfficacyEval/tree/main/data/SocialCD-3k}
          \item[2]  \url{https://github.com/YinpeiDai/NAUM}
          \item[3]  \url{https://github.com/405200144/Dataset-of-Cognitive-Distortion-detection-and-Positive-Reconstruction/tree/main}
          \item[4]  \url{https://www.kaggle.com/datasets/sagarikashreevastava/cognitive-distortion-detetction-dataset}
          \item[5] \url{https://huggingface.co/datasets/Psychotherapy-LLM/CBT-Bench}
          \item[6]  \url{https://github.com/facebookresearch/ParlAI/tree/main/projects/reframe_thoughts}
          \item[7]  \url{https://github.com/bcwangavailable/C2D2-Cognitive-Distortion}
          \item[8]  \url{https://github.com/behavioral-data/Cognitive-Reframing}
          \item[9]  \url{https://huggingface.co/datasets/halilbabacan/cognitive_distortions_gpt4}
          \item[10]  \url{https://osf.io/rsy4z/?view_only=41dc962f0c924c0e87e7bfc044535bd3}
          \item[11] \url{https://github.com/cocoboldongle/KoACD}
          \item[12] \url{https://huggingface.co/datasets/halilbabacan/combined_synthetic_cognitive_distortions}
          \item[13] \url{https://www.iitp.ac.in/~ai-nlp-ml/resources.html#DeCoDE-CoDEC}
          \item[14] \url{https://github.com/clang1234/ZS-CoDR.git}
        \end{tablenotes}
    
        \label{tab:cd_datasets_full}
      \end{threeparttable}
      \end{adjustbox}
      
    \end{table*}

}

\clearpage

\end{document}